\documentclass[conference,compsoc]{IEEEtran}
\usepackage{graphicx}
\usepackage{fancyhdr} 
\ifCLASSOPTIONcompsoc
  \usepackage[nocompress]{cite}
\else
  \usepackage{cite}
\fi
\ifCLASSINFOpdf
\else
\fi
\begin{document}
\title{Multimodal Sentiment Analysis Based on BERT and ResNet}

\author{\IEEEauthorblockN{JiaLe Ren}
\IEEEauthorblockA{School of Information and Engineering \\
Zhongnan University of Economics and Law  \\
Email: 3045597070@qq.com}
}

\maketitle

\begin{abstract}
With the rapid development of the Internet and social media, multi-modal data (text and image) is increasingly important in sentiment analysis tasks. However, the existing methods are difficult to effectively fuse text and image features, which limits the accuracy of analysis. 
To solve this problem, a multimodal sentiment analysis framework combining BERT and ResNet was proposed. BERT has shown strong text representation ability in natural language processing, and ResNet has excellent image feature extraction performance in the field of computer vision. Firstly, BERT is used to extract the text feature vector, 
and ResNet is used to extract the image feature representation. Then, a variety of feature fusion strategies are explored, and finally the fusion model based on attention mechanism is selected to make full use of the complementary information between text and image. Experimental results on the public dataset MAVA-single show that compared with the single-modal models that only use BERT or ResNet, 
the proposed multi-modal model improves the accuracy and F1 score, reaching the best accuracy of 74.5\%. This study not only provides new ideas and methods for multimodal sentiment analysis, but also demonstrates the application potential of BERT and ResNet in cross-domain fusion. In the future, more advanced feature fusion techniques and optimization strategies will be explored to further improve the accuracy and generalization ability of multimodal sentiment analysis.
\end{abstract}


%
\IEEEpeerreviewmaketitle

\section{Introduction}
\noindent As an important research direction in the field of natural language processing and artificial intelligence, sentiment analysis aims to identify, extract and classify emotional tendencies in text or multimedia data \cite{liu2022sentiment}. 
With the rise of social media and online platforms, sentiment analysis plays an increasingly important role in brand reputation management, user behavior analysis, and social event monitoring \cite{devika2016sentiment}. However, 
traditional sentiment analysis methods mainly rely on a single text or image data, ignoring the complementarity and correlation between multi-modal data, which limits the accuracy and comprehensiveness of the analysis results.
\\
\\
In order to overcome this limitation, multimodal sentiment analysis has gradually become a research hotspot in recent years. By fusing information from different modalities (such as text, image, audio, etc.), 
multimodal sentiment analysis can more accurately capture and understand the emotional expression of users. However, multi-modal sentiment analysis also faces many challenges, the most critical of which is how to effectively fuse features from different modalities, 
and how to maintain the integrity and consistency of information in the fusion process.
\\
\\
In the field of natural language processing, google published BERT (\textbf{B}idirectional \textbf{E}ncoder \textbf{R}epresentations from \textbf{T}ransformers) \cite{devlin2018bert} \cite{vaswani2017attention} pre-trained model in 2018. Due to its powerful text representation ability and context understanding ability, It has achieved remarkable results in several natural language processing tasks. 
Through large-scale pre-training and bidirectional Transformer architecture, BERT can capture the deep semantic information in the text, which provides a powerful tool for sentiment analysis. However, when dealing with image data, relying on bert alone will not achieve the best results
\\
\\
In contrast, \textbf{ResNet} (Residual Network) \cite{he2016deep} model has excellent performance in the field of computer vision. ResNet achieves deep extraction and efficient representation of image features by introducing residual connections and deep convolutional neural networks. 
However, ResNet also has limitations when dealing with text data.
\\
\\
Therefore, this paper proposes a multimodal sentiment analysis framework combining BERT and ResNet \cite{bhat2022exploring} \cite{anggrainingsih2022evaluating} \cite{udayakumar2023automatic}, which aims to utilize the advantages of BERT in text processing and ResNet in image processing to achieve effective multimodal data fusion and sentiment analysis. 
Through different model fusion methods, we can extract deeper comprehensive features. 
Through this framework, we expect to improve the accuracy of multimodal sentiment analysis and provide new ideas and methods for research and application in related fields.
\section{Related Work}
\subsection{Text sentiment analysis}

\begin{figure*}[!t]
  \centering
  \caption{\bf{BERT Model}}

  \includegraphics[width=12cm]{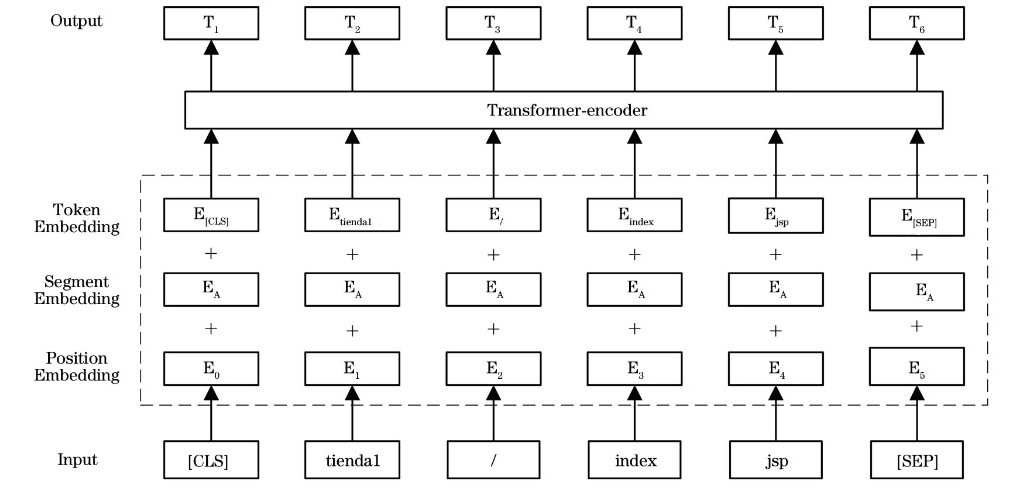}
\end{figure*}

\begin{figure*}[htbp]
  \centering
  \caption{\bf{ResNet 50}}
  \includegraphics[width=12cm]{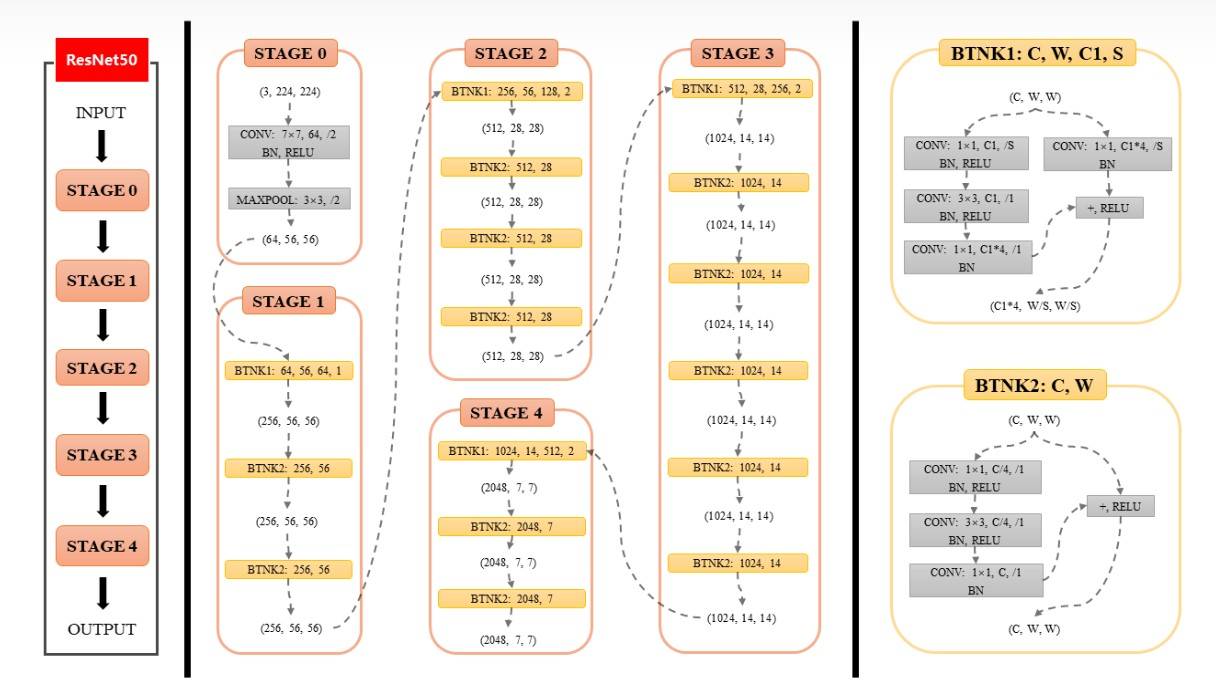}
\end{figure*}

\noindent Text sentiment analysis \cite{yadollahi2017current} is one of the core tasks in the field of natural language processing, which aims at automatically judging the emotional tendency of text expression through algorithms. 
Traditionally, sentiment analysis mainly relies on two approaches: rule-based and machine learning-based approaches.
\\ \\
Rule-based approaches usually rely on human-defined vocabularies and grammar rules to identify sentiment words and phrases in texts. Although this method may achieve certain results in a specific domain, it lacks flexibility and scalability, 
and is difficult to adapt to the changes of sentiment expression in different contexts and domains. In addition, manual definition of rules requires a lot of manpower and time, and it is difficult to comprehensively cover all possible emotional expressions.
\\ \\
Machine learning-based methods
With the development of machine learning technology, researchers begin to use machine learning algorithms such as Support vector Machine (SVM) \cite{platt1998sequential} and Naive Bayes \cite{dai2007transferring} \cite{laplace1774memoire} for sentiment analysis. These methods train models to automatically learn the mapping relationship between text features and sentiment labels, thus improving the accuracy of sentiment analysis to some extent. 
However, traditional machine learning algorithms still rely on manually designed feature engineering, which limits the generalization ability of the model. Moreover, these algorithms may face computational efficiency challenges when dealing with large-scale text data.
\\ \\
In order to solve the shortcomings of traditional sentiment analysis methods, deep learning technology such as CNN \cite{lu2021review} \cite{hershey2017cnn} RNN \cite{elman1990finding} LSTM \cite{hochreiter1997long}, especially pre-trained language models, has made significant progress in the field of NLP in recent years.
BERT(Bidirectional Encoder Representations from Transformers) model, as a pre-trained language model, 
has made remarkable achievements in the field of text sentiment analysis \cite{chiorrini2021emotion} \cite{sun2019fine}. BERT model is based on Transformer architecture and adopts bidirectional encoding to capture text context information more accurately. Its pre-training 
tasks include Masked Language Model (MLM) and Next Sentence Prediction (NSP). In the MLM task, the model randomly masks some words in the input text and predicts these words based on the context. The NSP task determines whether two sentences are contiguous text.
\\
\\
For an \textbf{MLM} task, given a text sequence of N terms $(w_{1},w_{2},...,w_{N})$ ,Where some words are obscured, the goal of the BERT model is to learn a conditional probability distribution as follows:

\begin{small}
$$P(w_{i}|w_{ !=i }) = softmax(W_{o}^{T} \cdot  BERT(w_{1},w_{2},...,[MASK],...,w_{N}))$$
\end{small}
\\

\noindent Where [MASK] represents the position of the masked word, $BERT(w_{1},w_{2},...,[MASK],...,w_{N})$ represents an encoded representation of the input text sequence by the BERT model.
\\
\\
For the \textbf{NSP} task, given two sentences A and B, the goal of the BERT model is to determine whether they are continuous text, i.e.

\begin{small}
$${P(IsNext|A,B) = \sigma (W_{c}^{t}\cdot [BERT(A);BERT(B)] )}$$
\end{small}
\\
Where $\bf{\sigma}$ is the sigmoid function,$\bf{W_{c}^{t}}$ is the weight matrix of the classifier, and [BERT(A);BERT(B)] represents the concatenation of the BERT encoded representations of the sentences A and B.
\\
\\
In text sentiment analysis, BERT model is usually used as a feature extractor, which input text into BERT to get its vector representation, and then use these features for emotion classification. BERT model can capture complex semantic relationships in text 
and improve the accuracy of sentiment analysis.

\subsection{Picture sentiment analysis}
\noindent Image emotion analysis \cite{zhao2014exploring} \cite{you2016building} is one of the important tasks in the field of computer vision, which aims to judge the emotional tendency of images by analyzing their content. As a kind of deep convolutional neural Network, ResNet (Residual Network) model 
solves the degradation problem in deep network training by introducing residual learning and residual block, and performs well in image sentiment analysis \cite{tian2022resnet} \cite{li2021facial}.
\\
\\
The core of the ResNet model is the residuals block, whose structure enables the network to learn the residuals between inputs and outputs directly, avoiding the problems of disappearing gradients and exploding gradients
\\
\\
For the $l$ th residual block, the input is $x_{l}$, the output is $x_{l+1}$, the residual function is $F(x_{l},W_{l})$ 
then the output of the residual block can be expressed as:
$$x_{l+1} = F(x_{l},W_{l}) + x_{l}$$
Where $F(x_{l},W_{l})$ represents the residual function, usually consisting of a convolution layer, a batch normalization layer, and a ReLU activation function, $W_{l}$ is the weight parameter of the residual function.
\\
\\
In image sentiment analysis, the ResNet model forms a deep network by stacking multiple residual blocks to extract image features. Then, the full connection layer and softmax function are used to classify the extracted 
features and judge the emotional tendency of the image expression.  The classification layer of the ResNet model is represented as follows:
\\
\\
Given the extracted image feature vector $f$, the classifier output of the ResNet model is:
$$ y = softmax(W_{f}^{T} \cdot f + b_{f})$$
Where, $W_{f}$ is the weight matrix of the classification layer, $b_{f}$ is the biased term, and $y$ is the probability distribution 
of the classification result.
\\
\\
\noindent ResNet model has strong feature extraction ability and good generalization performance, which can achieve excellent performance in image sentiment analysis tasks in different fields. In addition, the ResNet model is easy 
to train and converges quickly. By introducing regularization techniques such as batch normalization and discard layer, ResNet model can further improve its stability and robustness, which makes it have a wide application prospect 
in image sentiment analysis tasks.
\\
\\
To sum up, BERT and ResNet are representative models in text sentiment analysis and picture sentiment analysis respectively. They capture deep features in text and images through deep learning techniques, providing powerful 
support for sentiment analysis tasks. In future studies, the fusion methods of BERT and ResNet can be further explored, as well as their application to multimodal sentiment analysis tasks \cite{gandhi2023multimodal} \cite{han2021improving}.

\section{METHOD}
\noindent Compared with a single model, such as using BERT and ResNet for prediction, we use a multimodal model. Therefore, in this section, we will explain in detail 5 different multimodal models based on the fusion of BERT and ResNet50. 
Through different fusion methods, the model will learn different feature information.
\\
\\
The purpose of image sentiment analysis is to predict the sentiment tendency of users based on the images and texts they post. 
So the text and image inputs are as follows:
\\
\\
\textbf{Text input} : denoted $T = \{t_{1},t_{2},...,t_{N}\}$, where $t_{i}$ represents the i th text sample, where N is the number of the text sample.
\\
\\
\textbf{Image input} : denoted $I = \{i_{1},i_{2},...,i_{N}\}$, where $i_{j}$ represents the j th image sample.
\\
\\
\textbf{Feature extraction} :  $$H_{T}, F_{T} = BERT(T) $$ $$H_{I}, F_{I} = ResNet50(I) $$
where $H_{T}$ is the hidden state matrix of the text, $F_{T}$ is the eigenvector matrix of the text. 
$H_{I}$ is the hidden state matrix of the image, $F_{I}$ is the  eigenvector matrix of the image.

\subsection{CMACModel}
\begin{figure*}[htbp]
  \centering
  \caption{CMACModel}
  \includegraphics[width=12cm]{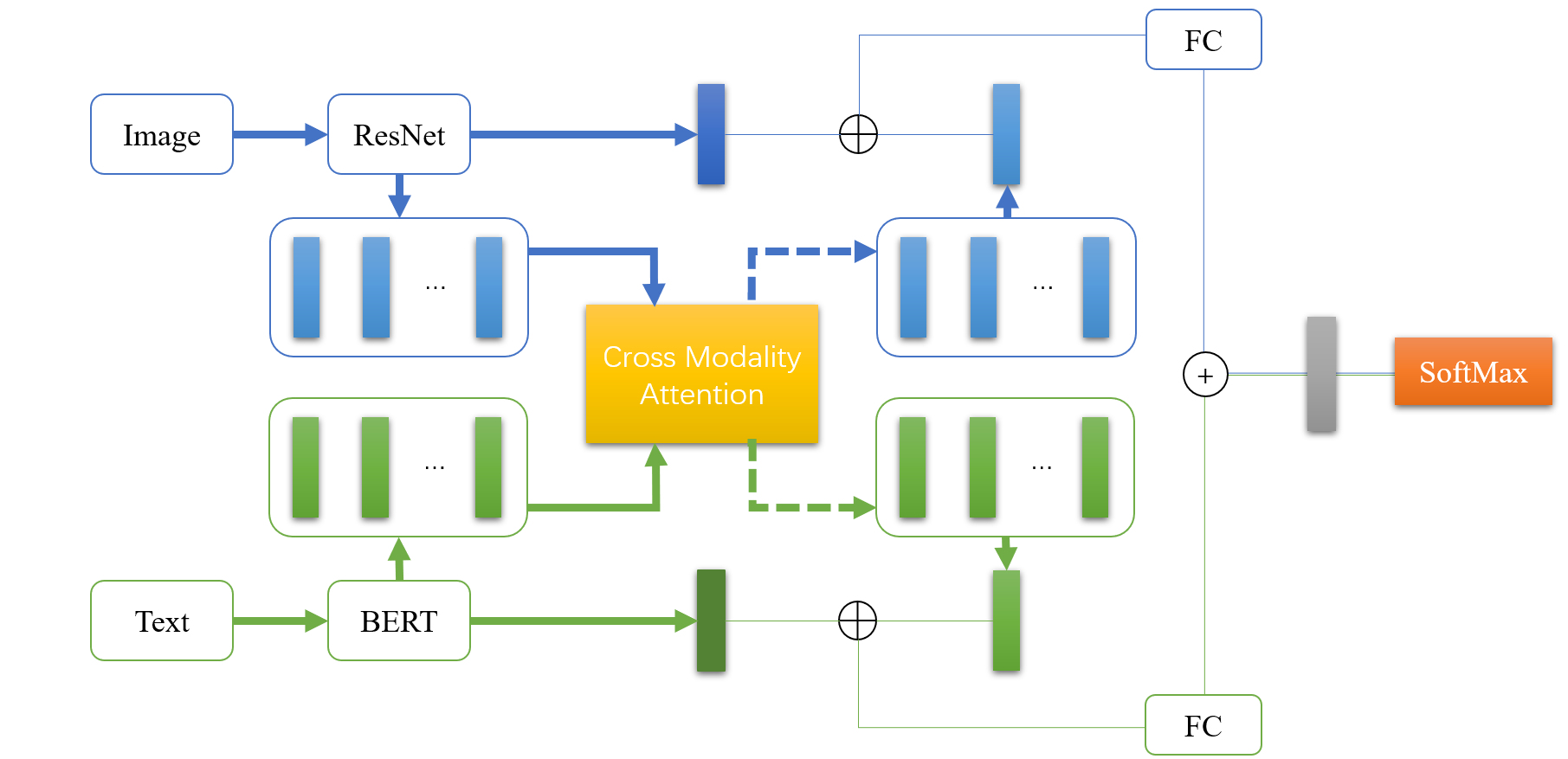}
\end{figure*}
\begin{figure*}[htbp]
  \centering
  \caption{HSTECModel}
  \includegraphics[width=12cm]{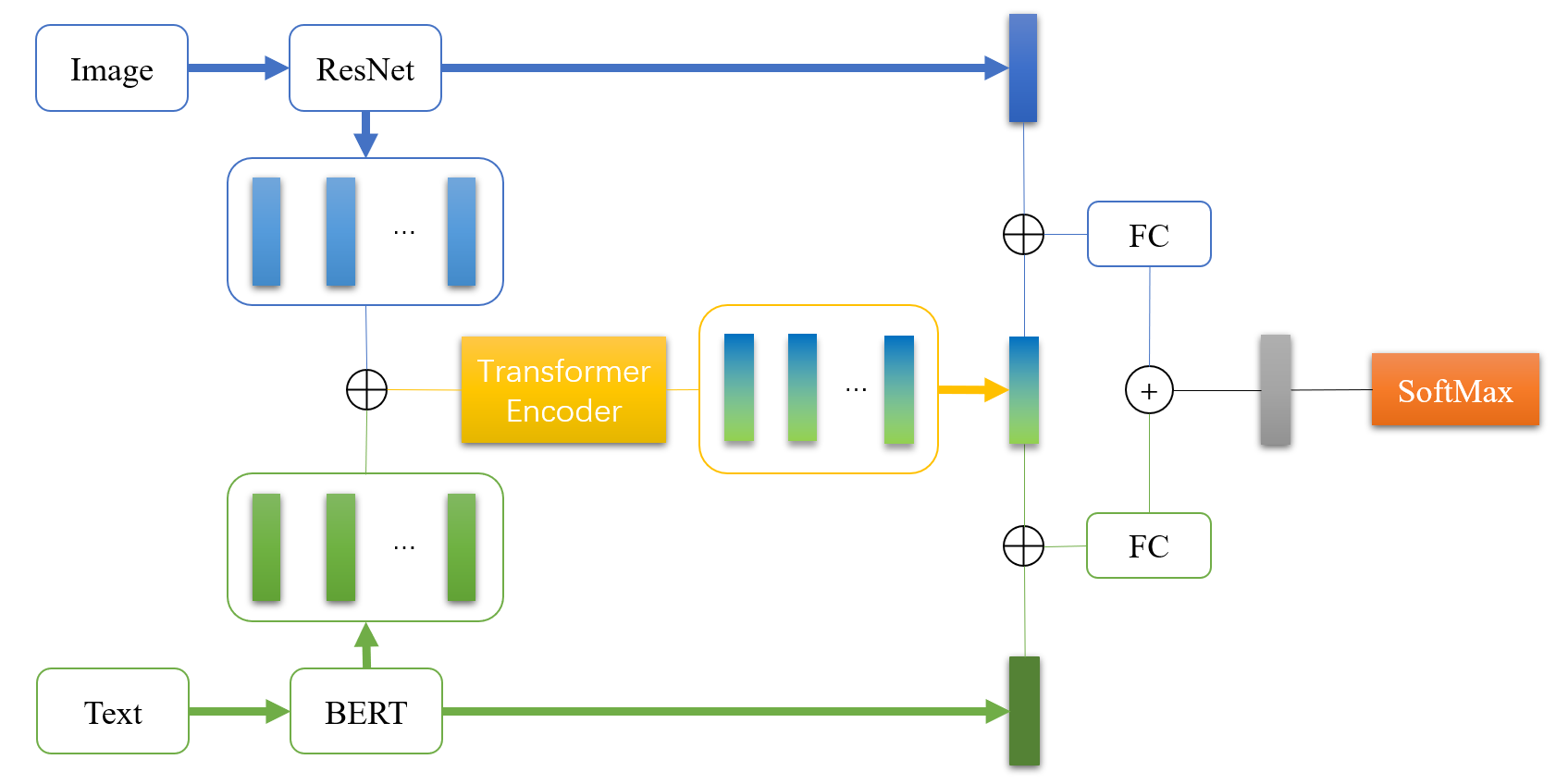}
\end{figure*}
\begin{figure*}[htbp]
  \centering
  \caption{OTEModel}
  \includegraphics[width=12cm]{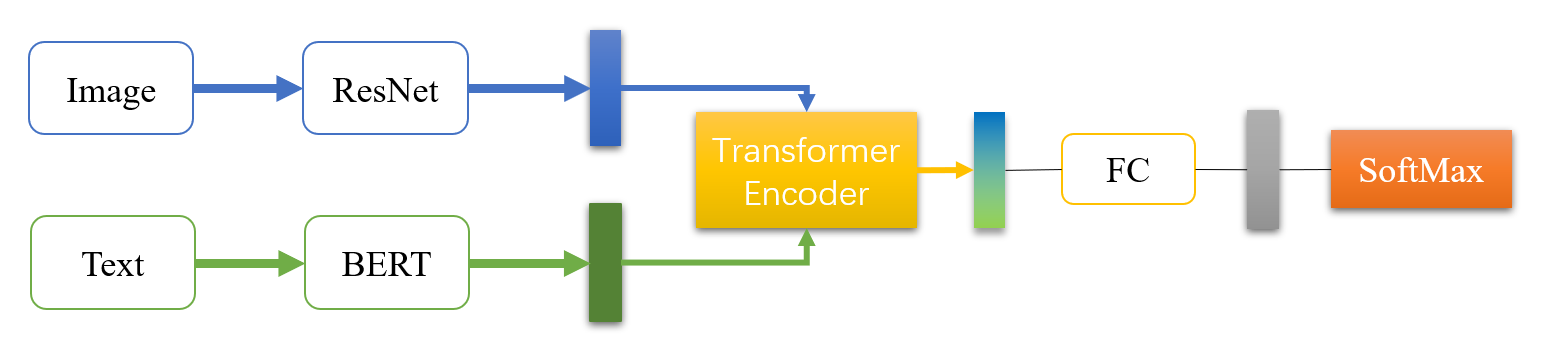}
\end{figure*}
\noindent  In this section we introduce \textbf{CMACModel} (CrossModalityAttentionCombineModel) \cite{xu2020cross} \cite{ye2019cross}. The model allows text and images to "understand" each other and fuse each other's information through a bidirectional, cross-modal attention mechanism. 
This two-way interaction allows the model to more accurately capture the associations between text and images, which improves the performance of the classification task.
\\
\\
\textbf{Cross Modality Attention}: $$A_{I->T} = Attention_{I->T}(H_{I}, H_{T}, H_{T})$$
$$A_{T->I} = Attention_{T->I}(H_{T},H_{I},H_{I})$$
where $A_{I->T}$ is the image-to-text attention output matrix. Then $A_{I->T}$ is averaged and compressed to get the attention output
vector $a_{I->T}$, By the same token, we get vector $a_{T->I}$.
\\
\\
\textbf{Classifier input and output}: $$p_{T} = TextClassifer([F_{T};a_{I->T}])$$ $$p_{I} = ImageClassifer([F_{I};a_{T->I}]) $$
where $[;]$ represents a vector concatenation operation, $p_{T}$ is the probability vector output by the text classifier. 
$p_{I}$ is the probability vector of the output of the image classifier
\\
\\
\textbf{Probability vector fusion and label prediction}:
$$p = softmax(p_{T} + p_{I})$$ $$y = argmax(p)$$

\subsection{HSTECModel}

\noindent In this section we introduce \textbf{HSTECModel} (HiddenStateTransformerEncoderCombineModel) \cite{aken2020visbert}\cite{hong2019state}by extracting the features of text and image, 
the model uses attention mechanism to fuse cross-modal information, and then makes classification prediction of text and image based on the fused information, and finally obtains the final prediction label 
by combining the prediction results of both.
\\
\\
\textbf{Attention computation}:
$$H_{concat} = [H_{T} ; H_{I}]$$ $$A = Attention(H_{concat})$$ $$a = mean(A,dim=0).squeeze(0)$$
where $\bf{a}$ is the attention output vector after averaging and squeezing operations.
\\
\\
\textbf{Classifier input and output:}
$$p_{T} = TextClassifer([F_{T};a])$$ $$ p_{I} = ImageClassifer([F_{I};a])$$
\\
\textbf{{Probability vector fusion and label prediction}} :
$$p = softmax(p_{T}+p_{I})$$
$$y = argmax(p,dim=1)$$

\subsection{OTEModel}

\noindent In this section, we introduce \textbf{OTEModel} (OutputTransformerEncoderModel) \cite{mehdy2021multi} \cite{miyairi1986new}, The model extracts the features of text and image, splices and fuses these features in the feature space to form a joint feature 
representation. This joint feature representation is then used in the classification task to predict the final label.The reasoning process is as follows:
$$F_{concat} = concat([F_{I},F_{T}],dim=2)$$
$$F_{attn} = Attention(F_{concat})$$
$$p = Classifier(F_{attn})$$
$$y = argmax(p,dim = 1)$$

\subsection{NativeModel}
\noindent We have two native combine model: \textbf{NativeCatModel} and \textbf{NativeCombineModel}
\\
\\
\textbf{NativeCatModel}: The model extracts the features of text and image respectively, and concatenates these features to form a joint feature vector. 
This joint feature vector is then used for a classification task to predict the final label
$$p = classifier([{F_{T};F_{I}}])$$
$$y = argmax(p,dim=1)$$
\\
\\
\textbf{NativeCombineModel}: The model processes text and image input separately, extracts their respective features, and makes classification predictions independently. 
Then, the prediction probability of text and image is fused to get the final prediction result.
$$p_{T} = TextClassifer(F_{T})$$
$$p_{I} = ImageClassifer(F_{I})$$
$$p = softmax((p_{T}+p_{I}), dim = 1)$$
$$y = argmax(p,dim=1)$$

\begin{table*}[htbp]
  \begin{center}
   Table1 datasets  
  \setlength\tabcolsep{20pt}
\renewcommand{\arraystretch}{1.4}
\begin{tabular}{c c c}
\hline
Image     &   Text     &   Labels \\ \hline
\\
\begin{minipage}[b]{0.4\columnwidth}
\centering
\raisebox{-.5\height}{\includegraphics[width=\linewidth]{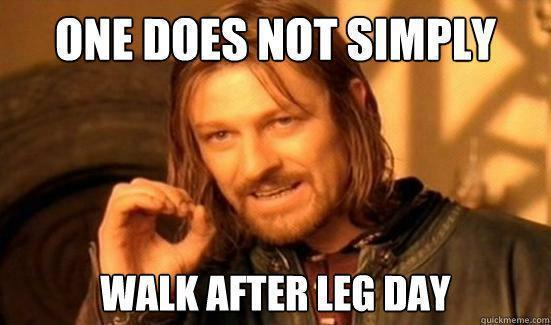}}
\end{minipage}
  &    How I feel today \#legday \#jelly \#aching \#gym   
  &    positive  
  \\ \\ \hline  
\\ 
\begin{minipage}[b]{0.4\columnwidth}
  \centering
  \raisebox{-.5\height}{\includegraphics[width=\linewidth]{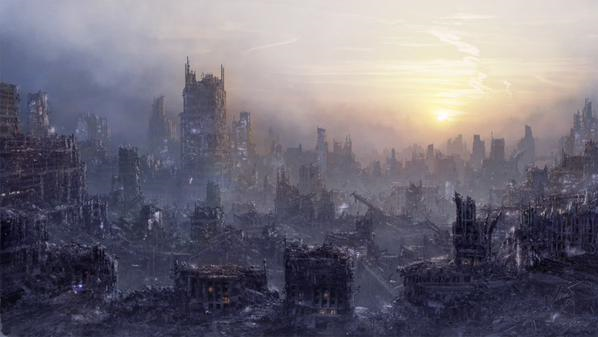}}
\end{minipage}   
    & "@NYSE is looking a little despondent today...???
    &   negative     \\ \\ \hline 
\\
\begin{minipage}[b]{0.3\columnwidth}
  \centering
  \raisebox{-.5\height}{\includegraphics[width=\linewidth]{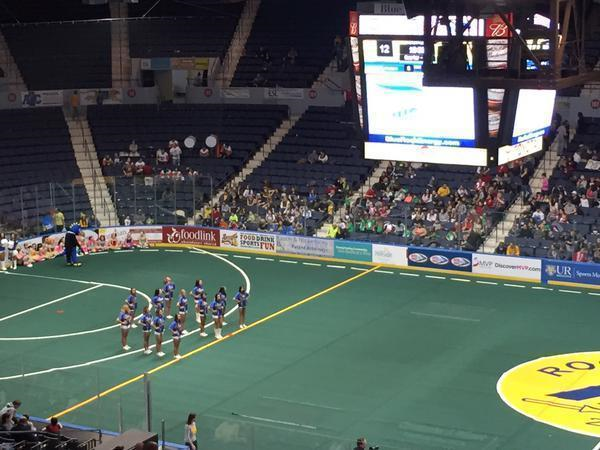}}
\end{minipage}     
&  Poor @HarrisburgHeat getting crushed by @RLancers 12 to 4 at half time.
&  neural     \\  
 \hline
\end{tabular}
    
\end{center}
\end{table*}

\section{EXPERIMENTS}

\subsection{datasets}
\noindent Experiments are carried out on a multi-modal public dataset MAVA-single
Crawled from the official twitter website, as shown Table 1, Dataset -
4,512 pairs of image-text data are included, each labeled with a correspondence
Positive emotions such as positive, negative, and neutral emotions
We split the dataset into Train, Val, Test as follows
Table 2

\begin{table}[htbp]
\begin{center}
    
\setlength\tabcolsep{40pt}
Table 2
    
\renewcommand{\arraystretch}{1.4}
\begin{tabular}{c c}
\hline
Data partitioning     & quantity             \\ \hline
Train &    3200     \\  
Val    &   800      \\  
Test    &  512        \\  
 \hline
\end{tabular}
    
\end{center}
\end{table}

\subsection{Experimental details}
\subsubsection{Hyperparameter configuration}
\noindent Parameter configuration can help us better configure the network structure, and then modify the network to achieve better results. 
The following figure is some configuration of the network hyperparameters, including bert\_embedding, num\_header, and so on

\begin{table}[htbp]
  \begin{center}
          
  \setlength\tabcolsep{30pt}
  Table 3
          
  \renewcommand{\arraystretch}{1.4}
  \begin{tabular}{c c}
  \hline
  parameter     &    Parameter value             \\ \hline
  bert\_embedding &   768     \\  
  num\_header    &  12      \\  
  bert\_dropout  &  0.1         \\  
  batch\_size    &   16       \\
  learning\_rate    & 3e-5      \\
  epoch        &  20         \\
  weight\_decay    & 0           \\
  \hline
  \end{tabular}
          
  \end{center}
  \end{table}

\subsubsection{Evaluation metrics}

\noindent In order to comprehensively and accurately evaluate the performance of our proposed multimodal sentiment analysis framework combining BERT with ResNet, we choose the following four key evaluation metrics: 
\textbf{Accuracy, Precision, Recall}, and \textbf{F1 Score}. These metrics can reflect the performance of the model from different perspectives and help us understand the strengths and weaknesses of the model more deeply.
$$Acc = \frac{TP+TN}{TP+TN+FP+FN}$$
$$Pre = \frac{TP}{TP+FP}$$
$$Recall = \frac{TP}{TP+FN}$$
$$F1\_score = \frac{2\cdot{Pre}\cdot{Recall}}{Pre+Recall}$$
\\
\subsection{Contrast experiment}

\begin{table*}[htbp]
  \begin{center}
          
  \setlength\tabcolsep{20pt}
  Table 4
          
  \renewcommand{\arraystretch}{1.4}

  \begin{tabular}{c c c c c}
  \hline
  Model     &    Acc/\%  &Pre/\%   & Recall/\%  &F1 score/\%   \\ \hline
  Bert      &     70.5   & 71    &    70     &     70  \\
  ResNet    &    65.75    &  65  &   66    &   65     \\ \hline
  CMACModel &   73.13 &      66 &      73  &     69    \\  
  HSTECModel    &  71.75  &  64 &       72  &     67  \\  
  \textbf{OTEModel}  &  \bf{74.5}    &     \bf{73} &    \bf{74}   &  \bf{73}        \\  
  NativeCatModel   &70.5 &    70   &    70  &    70    \\
  NativeCombineModel    & 71.5  &71   &  71 &   71       \\
  \hline
  \end{tabular}
              
  \end{center}
  \end{table*}

\noindent In order to more deeply understand the performance advantages of the multimodal sentiment analysis framework combining BERT and ResNet, and verify its applicability in different contexts, we designed a series of comparative experiments. 
These trials are designed to fully evaluate the performance of our proposed model by comparing the performance of different models, different combinations of features, and different datasets. The statistics of the results of the experiments are shown in Table4.

\subsection{Results Analysis}
\noindent From the above experimental results, it can be seen that the best performance on the dataset is achieved by using the \textbf{OTEModel}.
\\
\\
\textbf{Acc} : Our proposed OTEModel model achieves 4\% and 8.75\% improvement in accuracy over the single-modal BERT model and ResNet model, 
respectively, and 4\% improvement over the simple multimodal fusion model. This indicates that our model has better classification ability on the whole and can identify sentiment tendencies more accurately.
\\
\\
\textbf{Pre} : In terms of Precision, our OTEModel model outperforms the single-modal BERT model and ResNet model by 2\% and 8\%, respectively, 
and outperforms the simple multimodal fusion model by 3\%. A high precision means that our model makes fewer false positives when predicting the positive class, which is crucial for sentiment analysis tasks in real-world applications.
\\
\\
\textbf{Recall} : Recall results show that our OTEModel model improves by 4\% and 8\% over the single-modal BERT model and ResNet model, respectively, 
and improves by 4\% over the simple multimodal fusion model. This indicates that our model is able to identify more true positive samples and improve the integrity of sentiment analysis.
\\ \\
\textbf{F1 Score} : Our OTEModel model also achieves a significant advantage in F1 score, achieving 3\% and 8\% improvement over the single-modal BERT model and ResNet model respectively, 
and 3\% improvement over the simple multi-modal fusion model. The improvement of F1 score further verifies that our model achieves a good balance between precision and recall and has stronger comprehensive performance.
\\ \\
Our experimental results show that the multimodal sentiment analysis framework combining BERT with ResNet can make full use of the complementary information between text and image and improve the accuracy of sentiment analysis. Compared with the single-modal model, the multimodal model can capture more emotional cues and thus judge the emotional tendency more accurately. 
In addition, compared with the simple multimodal fusion model, our model further improves the performance of sentiment analysis by introducing more complex feature fusion strategies, such as attention mechanism.
\\ \\
However, we also note that although our model achieves a significant advantage in overall performance, the performance of the model still needs to be improved in some specific cases, such as extreme emotional tendencies or complex background images. Therefore, in future research, 
we will continue to explore more advanced feature extraction and fusion techniques to further improve the performance of the model.

\section{Conclusion}
\noindent This study proposes a sentiment analysis method based on the fusion of BERT and ResNet50 models. We explored five different model fusion methods, aiming to combine the text understanding ability of BERT and the visual feature extraction ability 
of ResNet50 to improve the accuracy and robustness of sentiment analysis. Experimental results show that OTEModel (\textbf{O}utput\textbf{T}ransformer\textbf{E}ncoder\textbf{Model}) performs best among all five methods and successfully achieves the best sentiment 
classification effect on standard datasets. Specifically, the BERT model provides accurate text representation for sentiment analysis through its powerful language understanding ability, while ResNet50 plays an important role in processing sentiment 
analysis tasks containing visual information by extracting image features. We adopted a variety of strategies in the model fusion process, including weighted averaging, feature-level fusion, decision-level fusion, etc. Among them, OTEModel fully utilized the advantages of 
both in sentiment analysis tasks through effective feature integration and optimization strategies.
\\
\\
\noindent The contribution of this study is that the BERT and ResNet50 models are combined and applied to sentiment analysis tasks, and the innovative fusion method OTEModel is proposed, which provides new ideas for research and application in the field of sentiment analysis. 
Future work can further explore more complex multimodal fusion techniques and their application effects in practical scenarios.

\bibliography{bare_conf_compsoc} 
\bibliographystyle{ieeetr} 



\end{document}